\documentclass{article}

\usepackage[final]{neurips_2024}

\usepackage[utf8]{inputenc} 
\usepackage[T1]{fontenc}    
\usepackage{hyperref}       
\usepackage{url}            
\usepackage{booktabs}       
\usepackage{amsfonts}       
\usepackage{nicefrac}       
\usepackage{microtype}      
\usepackage{xcolor}         
\usepackage{graphicx}
\usepackage{multirow}
\usepackage{adjustbox}
\usepackage{array}
\usepackage{caption}
\usepackage{subcaption}
\usepackage{svg}
\usepackage{forest}
\usepackage{todonotes}
\usepackage{xcolor,colortbl}
\usepackage{amsmath,amsfonts,amssymb}
\definecolor{hlgreen}{HTML}{B2D5CB}
\definecolor{hlblue}{HTML}{ADD8E6}
\definecolor{hlyellow}{HTML}{EADDCA}

\usepackage{xspace}

\usepackage{amsmath}

\DeclareSymbolFont{extraup}{U}{zavm}{m}{n}
\DeclareMathSymbol{\varheart}{\mathalpha}{extraup}{86}

\title{MERaLiON-AudioLLM: Bridging Audio and Language with Large Language Models}

\author{%
  {\Large MERaLiON Team}
  \AND
  Yingxu He$^{\tiny *}$, Zhuohan Liu$^{\tiny *}$, Shuo Sun$^{\tiny *}$, Bin Wang$^{\tiny *}$, Wenyu Zhang$^{\tiny *}$, Xunlong Zou\thanks{Core contributors listed by alphabetical order. Please cite this report as authored by MERaLiON team. \\
  Corresponding: \{sun\_shuo, wang\_bin\}@i2r.a-star.edu.sg
  \\
  Available at: \url{https://huggingface.co/MERaLiON/MERaLiON-AudioLLM-Whisper-SEA-LION}
  }
  \vspace{2mm}\\
  \textbf{Nancy F. Chen, Ai Ti Aw} 
  \vspace{2mm}\\
  Institute for Infocomm Research (I$^2$R), A*STAR, Singapore
}

\begin{document}

\maketitle

\begin{abstract}
We introduce MERaLiON-AudioLLM (\textbf{M}ultimodal \textbf{E}mpathetic \textbf{R}easoning \textbf{a}nd \textbf{L}earning \textbf{i}n \textbf{O}ne \textbf{N}etwork), the first speech-text model tailored for Singapore’s multilingual and multicultural landscape. 
Developed under the National Large Language Models Funding Initiative, Singapore, MERaLiON-AudioLLM integrates advanced speech and text processing to address the diverse linguistic nuances of local accents and dialects, enhancing accessibility and usability in complex, multilingual environments. Our results demonstrate improvements in both speech recognition and task-specific understanding, positioning MERaLiON-AudioLLM as a pioneering solution for region-specific AI applications. We envision this release to set a precedent for future models designed to address localised linguistic and cultural contexts in a global framework.
\end{abstract}

\section{Introduction}

The rapid advancements in generative AI and large language models (LLMs) have unlocked transformative possibilities across various industries, particularly in domains requiring multimodal intelligence \citep{Minaee2024LargeLMsurvey, cui2024survey, ji2024wavchatsurvey}. Recognising the strategic importance of developing AI capabilities tailored to local needs, we initiated the development of the nation's first AudioLLM (Audio Large Language Model) under the National Large Language Models Funding Initiative, Singapore \citep{nllm}. 
This initiative aims to integrate advanced speech and text understanding, enabling seamless communication and nuanced comprehension of Singapore’s multilingual and multicultural landscape, while improving performance on region-specific benchmarks.

Our development journey began with establishing a robust infrastructure for multimodal LLM training, including a distributed data pipeline capable of processing over 30 TB of speech-text datasets and scalable training workflows deployed across high-performance H100 GPU clusters. To overcome the limitations of low-resource datasets, particularly in spoken question answering and dialogue summarization, we enhanced the data pipeline with synthesised and augmented datasets to ensure broader coverage of linguistic diversity.
Through iterative training and evaluation, our model demonstrates the ability to balance a 10-billion-parameter architecture while maintaining both computational efficiency and task accuracy.

In this technical report, we introduce the first version of MERaLiON-AudioLLM (\textbf{M}ultimodal \textbf{E}mpathetic \textbf{R}easoning \textbf{a}nd \textbf{L}earning \textbf{i}n \textbf{O}ne \textbf{N}etwork), a speech-text model designed with Singapore-specific linguistic and cultural adaptations. The development of a model that understands local accents and contextual nuances is crucial for creating more inclusive and effective AI systems. Traditional speech recognition models often struggle with the diversity of accents, dialects, and linguistic subtleties, leading to inaccuracies and reduced usability in complex, multilingual environments. By addressing these challenges, MERaLiON-AudioLLM aims to deliver robust performance across a wide range of speech patterns, improving accessibility and ensuring reliable communication for diverse populations.

This release marks the beginning of our mission to enhance Singapore's capabilities in multilingual and multimodal modeling. 
We will continuously refine our models and share our findings with the community, to drive progress and foster innovation.
Our model is publicly available at \url{https://huggingface.co/MERaLiON/AudioLLM}.

\section{Approach}
MERaLiON-AudioLLM is developed with a focus on enhancing its understanding of local accents and contextual nuances. 
In this initial release, our model is designed to take in a (audio, text) pair as input and generates a text output.
We carefully curated a diverse collection of datasets by combining real-world speech data with synthesised and augmented samples to improve representation across various accents and linguistic contexts. 
Using these curated datasets, we fine-tuned MERaLiON-Whisper encoder from Whisper-large-v2 \citep{radford2023whisper} and fused it with SEA-LION V3 \citep{sea_lion_2024}, a localised LLM developed by our partner AI Singapore,\footnote{\url{https://aisingapore.org/}} enabling integration of auditory and textual information in an end-to-end manner for downstream tasks. This offers better flexibility in decoding speed, multitask learning, and avoid error propagation in cascaded models.
In future iterations, we plan to extend the model’s context length and enhance its ability to handle multi-turn interactions and interleaved audio-text inputs. These improvements will expand its applicability in conversational and multimodal scenarios.

\subsection{Model Architecture}
\begin{figure}[tb]
    \centering
    \includegraphics[width=0.65\textwidth]{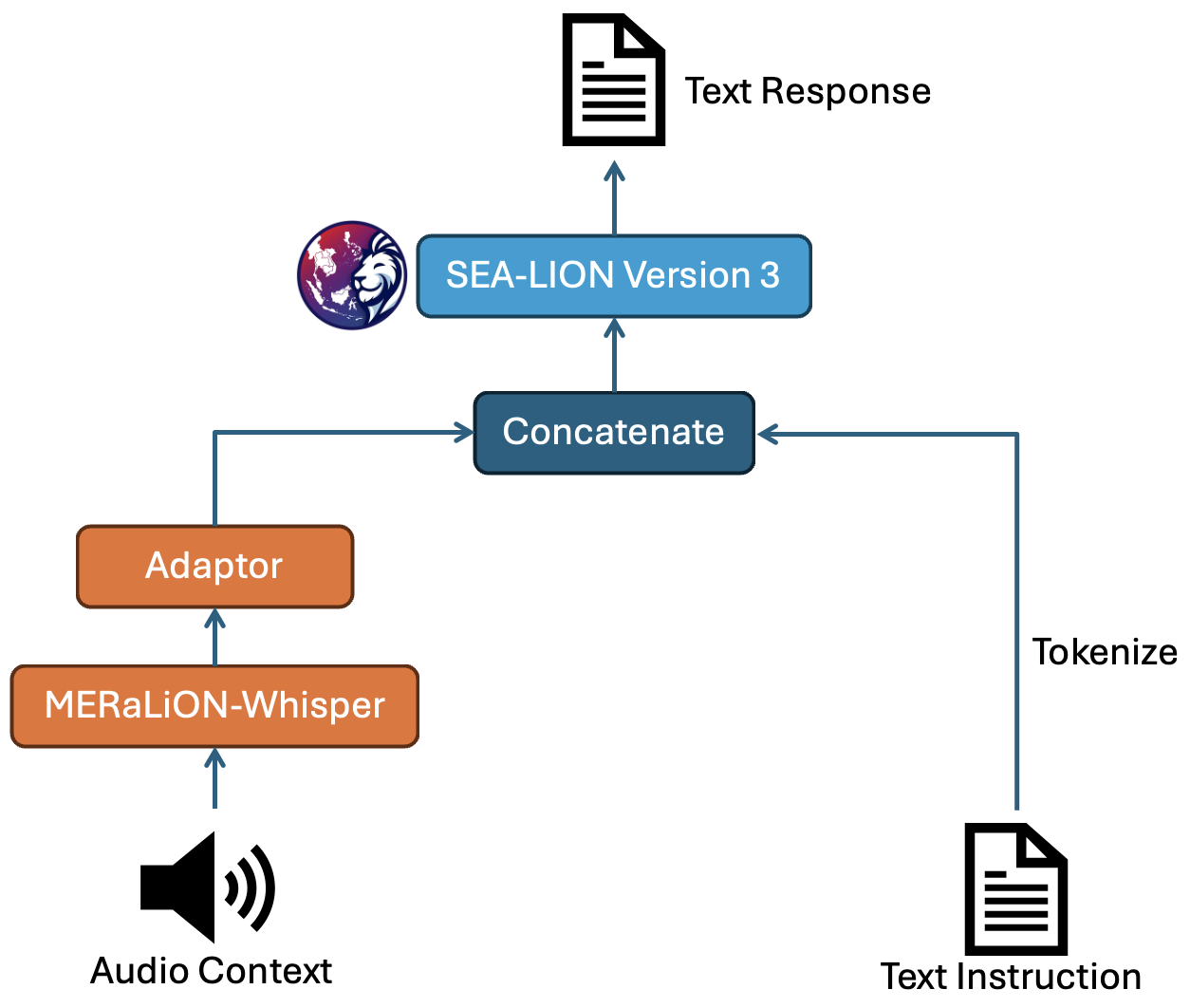}
    \caption{Architecture of \textbf{MERaLiON-AudioLLM}: MERaLiON-AudioLLM fuses MERaLiON-Whisper with AI Singapore's SEA-LION V3.}
    \label{fig:model architecture}
\end{figure}

MERaLiON-AudioLLM adopts a fusion-based architecture, following designs similar to other AudioLLMs \citep{fang2024LlamaOmni, defossez2024moshi, gong2024ltu, ghosh2024GAMA, chu2023qwenaudio, Qwen2-Audio}. 
The architecture, illustrated in Figure~\ref{fig:model architecture}, comprises three key components: an audio encoder that transforms speech or audio inputs into sequences of vector representations, a text decoder that interprets and responds to natural language instructions, and an adaptor module that compresses the encoder representations while aligning the encoder's hidden dimension with the text decoder's embedding size.

\subsubsection{Audio Encoder}
The purpose of the audio encoder is to transform an input speech or audio into a sequence of vector representations.
Formally, given an input sequence of raw audio signals $\mathbf{x} = \{x^{(1)}, x^{(2)}, \dots, x^{(T)}\}$ and an encoder $\phi$, the encoder maps $\mathbf{x}$ to a sequence of hidden representations $\mathbf{h}$:
\[
\mathbf{h} = \phi(\mathbf{x}), \mathbf{h} \in \mathbb{R}^{\tau \times d}
\]
where $\tau$ represents the output sequence length, $d$ is the hidden size, and $\tau \ll T$.
In this work, we build on the encoder of Whisper-large-v2 \citep{radford2023whisper}, which has demonstrated strong performance across various speech recognition tasks, to develop our in-house MERaLiON-Whisper.
Whisper processes audio sampled at 16,000 Hz by converting it into log-Mel spectrogram representations.
To adapt Whisper to local accents and linguistic contexts, we further fine-tune the model using a mixture of publicly available and in-house automatic speech recognition (ASR) datasets.

We are also exploring the integration of a localised speech encoder, which has been pre-trained from scratch using a self-supervised learning (SSL) framework, and will introduce it in future work~\cite{huzaifah2024speechfoundationmodelsingapore}.


\subsubsection{Adaptor Module}

As the encoder's hidden dimension is significantly smaller than the embedding size of the text decoder, we employ an adaptor module to align the speech or audio embeddings with the text embedding space. Specifically, MERaLiON-Whisper produces embeddings of sequence length 1500 and hidden dimension 1280, while SEA-LION V3 has an embedding size of 3854. We utilize a simple yet effective multi-layer perceptron (MLP) adaptor module with 2 hidden layers to transform the encoder outputs into 100 speech or audio token embeddings with a dimensional size of 3854. We refer to this simple adaptor module as \textbf{MLP-100}. It yields slightly better results compared to other alternatives such as window-level Qformer \citep{tang2024salmonn} and ConvMLP \citep{li2023convmlp}.

Formally, given an encoder output sequence $\mathbf{h} \in \mathbb{R}^{\tau \times d}$, the adaptor module first reshapes the sequence to $\tilde{\mathbf{h}} \in \mathbb{R}^{(\tau/s) \times (d \cdot s)}$, where $s$ is a scaling factor set to 15.
This operation effectively reduces the sequence length of $\mathbf{h}$ by concatenating outputs from $s$ time steps.
The adaptor then applies a linear transformation to $\tilde{\mathbf{h}}$:
\[
\mathbf{z} = \text{SiLU}(\tilde{\mathbf{h}} \mathbf{W_1}^\intercal + \mathbf{1}_{(\tau/s)} \mathbf{b_1}^\intercal)
\]

where $\mathbf{W_1} \in \mathbb{R}^{d \times (d \cdot s)}$ and $\mathbf{b_1} \in \mathbb{R}^{d}$ are learnable parameters in the first MLP layer, $\mathbf{1}_{(\tau/s)}$ is a vector of 1's of length $\tau/s$ used to broadcast the bias vector $\mathbf{b_1}$ into the shape of $\tilde{\mathbf{h}} \mathbf{W_1}^\intercal$, SiLU is the Sigmoid Linear Unit activation function \citep{hendrycks2023gaussianerrorlinearunits, elfwing2018sigmoid}, and the resulting representation is $\mathbf{z} \in \mathbb{R}^{(\tau/s) \times d}$.
This is followed by upward and downward projection layers:

\begin{equation*}
\begin{split}
    \mathbf{z_u} & =  \text{SiLU}(\mathbf{z}\mathbf{W_u}^\intercal + \mathbf{1}_{(\tau/s)} \mathbf{b_u}^\intercal) \\
    \mathbf{z_d} & = \mathbf{z_u}\mathbf{W_d}^\intercal + \mathbf{1}_{(\tau/s)} \mathbf{b_d}^\intercal
\end{split}
\end{equation*}

where $\gamma$ is the embedding size of the text decoder, $\mathbf{W_u}\in\mathbb{R}^{4d \times d}$, $\mathbf{b_u} \in \mathbb{R}^{4d}$, $\mathbf{W_d} \in \mathbb{R}^{\gamma \times 4d}$, and $\mathbf{b_d} \in \mathbb{R}^{\gamma}$ are trainable parameters. 
The final projected representation, $\mathbf{z_d}\in \mathbb{R}^{(\tau/s) \times \gamma}$, aligns the speech or audio embeddings into the same space as the text embeddings, allowing the model to process multimodal inputs jointly.

\subsubsection{Text Decoder}
The text decoder of MERaLiON-AudioLLM ingests a concatenated sequence of audio context tokens and text instruction tokens, and then generates a text-based response.
For this purpose, we leverage on SEA-LION V3 \citep{sea_lion_2024}, a state-of-the-art localised large language model developed by our partner, AI Singapore.
SEA-LION V3 was built upon the 9B version of Google's Gemma 2 \citep{gemmateam2024gemma2improvingopen} by continual pre-training it on an additional 200 billion tokens sourced from diverse datasets. These datasets encompass the four official languages of Singapore (English, Chinese, Malay, and Tamil), and also include several other Southeast Asian languages.
We use the instruct version of SEA-LION V3,\footnote{\url{https://huggingface.co/aisingapore/gemma2-9b-cpt-sea-lionv3-instruct}} which was further fine-tuned on approximately 500,000 English instruction-tuning pairs and approximately 1 million instruction tuning pairs in various ASEAN languages.

\subsection{Training Methodology}
\subsubsection{Audio Encoder Fine-tuning}
Whisper-large-v2 \citep{radford2023whisper} is an encoder-decoder ASR model by OpenAI that is well known for its robust performance in speech recognition across multiple languages.
Its ability to capture rich audio representations has made it a top choice for integration into AudioLLMs.
However, as a general-purpose ASR model, Whisper may not fully capture the subtle intricacies of local speech variations.
Therefore, we first guide the model to better capture these local characteristics by training it end-to-end on a collection of cleaned local ASR datasets derived from IMDA's National Speech Corpus, in-house ASR datasets and public datasets.

\subsubsection{Multimodal Instruction Fine-tuning}

We apply multimodal instruction fine-tuning to leverage datasets from multiple tasks delineated in Section~\ref{sec: tasks and datasets} to align MERaLiON-Whisper and SEA-LION V3 for cross-modal reasoning and task-specific performance. Let $\mathcal{D} = \bigcup_{j=1}^M \mathcal{D}_j$, where each $\mathcal{D}_j = \{(\mathbf{x}^{\text{audio}}_{i,j}, \mathbf{x}^{\text{text}}_{i,j}, \mathbf{y}_{i,j})\}_{i=1}^{N_j}$ represents dataset $j$ containing $N_j$ samples, with $\mathbf{x}^{\text{audio}}_{i,j}$ as speech or audio inputs, $\mathbf{x}^{\text{text}}_{i,j}$ as text inputs (such as task instructions or context), and $\mathbf{y}_{i,j}$ as task-specific outputs. 

To train MERaLiON-AudioLLM, we minimise the autoregressive loss function that measures the difference between the predicted and ground truth sequences. The model predicts for the output sequence $\mathbf{y}_{i,j} = \{\mathbf{y}_{i,j}^{(1)}, \mathbf{y}_{i,j}^{(2)}, \dots, \mathbf{y}_{i,j}^{(L)}\}$ autoregressively, where $L$ is the output sequence length. 
The autoregressive loss for a sample is formulated as:
\begin{equation}
    \mathcal{L}_{i,j} = \sum_{\ell=1}^{L} - \log P(\mathbf{y}_{i,j}^{(\ell)} \mid \mathbf{y}_{i,j}^{(<\ell)}, \mathbf{x}^{\text{audio}}_{i,j}, \mathbf{x}^{\text{text}}_{i,j})
\end{equation}
where $\mathbf{y}_{i,j}^{(<\ell)}$ represents the output tokens before the current prediction token.
This loss encourages the model to accurately predict each token in the output sequence, conditioned on the prior output tokens and the multimodal input representations. 

During training, we fully fine-tune the audio encoder and adaptor module, while partially fine-tuning the SEA-LION V3 text decoder by adding LoRA (Low-Rank Adaptation) \citep{hu2022lora} layers with a rank of 8 to all MLP layers.
We used the fused AdamW optimizer in PyTorch, along with a linear learning rate scheduler that includes 100 warm-up steps and a peak learning rate of 5e-5. To mitigate overfitting to artifacts in the input audio log-Mel spectrograms, we find it helpful to apply spectrogram augmentation \citep{park2019specaugment} by randomly masking a sequence of 20 time steps with a probability of 5\%.

\section{Datasets}
\label{sec: tasks and datasets}





We curated an extensive collection of speech-text instruction-tuning pairs totalling 260,000 hours of data. A significant portion of this dataset is derived from IMDA's National Speech Corpus (NSC) \citep{koh19_interspeech}, which is licensed under the Singapore Open Data License.\footnote{\url{https://data.gov.sg/open-data-licence}}
The National Speech Corpus contains approximately 10,600 hours of recordings of Singaporean English speakers, structured into six parts:
\begin{itemize}
    \item \textbf{Part 1} --- 3000 hours of prompted readings from phonetically balanced scripts
    \item \textbf{Part 2} --- 3000 hours of prompted readings featuring sentences on topics such as people, food, locations, and brands
    \item \textbf{Part 3} --- 900 hours of conversational data, including discussions on daily life and gameplay interactions
    \item \textbf{Part 4} --- 900 hours of code-switching conversations where speakers alternate between Singlish and their Mother Tongue languages (Chinese, Malay, Tamil).
    \item \textbf{Part 5} --- 1500 hours of conversations following four themes: debate, finance, positive emotion, and negative emotion.
    \item \textbf{Part 6} --- 1300 hours of simulated phone calls across three thematic designs: (1) holiday, hotel, restaurant, (2) bank, telephone, insurance, and (3) Housing and Development Board (HDB), Ministry of Education (MOE), Ministry of Social and Family Development (MSF).
\end{itemize}
Parts 1 to 5 also include detailed speaker meta-information such as gender, age, ethnic group, and first language, which we used to construct high-quality paralinguistics datasets. We also released the Multitask National Speech Corpus (MNSC) for open usage.\footnote{\url{https://huggingface.co/datasets/MERaLiON/MNSC}}

Although NSC serves as an invaluable resource for model training, it includes a significant amount of mislabelled data, as well as systematic and accidental errors. To ensure the integrity and reliability of the datasets, we performed thorough verification and filtering processes, extracting only the most accurate and high quality segments.
For Parts 1 and 2, we ensured that all examples with the same transcription were consistently assigned to the same data splits to avoid data leakage.
For Parts 3 to 6, we selected recordings where the audio duration closely matched the transcription timestamp duration. For conversational audio recorded separately for each speaker, we superimposed the speech from both sides by summing their respective audio array representations.
We further segmented longer conversations into shorter segments, each lasting up to 30 seconds.
For transcription notations, we removed non-speech content such as <mandarin>, <S>, and (ppb) while retaining discourse particles (e.g., [oh]), interjections (e.g., !walao!), and fillers (e.g., (um)).

We further expanded the dataset by synthesising examples for tasks such as Spoken Dialogue Summarization (SDS), Speech Question Answering (SQA), and Gender Recognition (GR). Detailed information on these datasets will be published separately.

\section{Technical Details}
\subsection{Compute and Infrastructure}
We are grateful for the invaluable support provided by the National Supercomputing Centre (NSCC), Singapore, which enabled us to complete our recent MERaLiON-AudioLLM training runs on the ASPIRE 2A+ Supercomputer.\footnote{\url{https://help.nscc.sg/aspire2aplus/about/}}
The ASPIRE 2A+ system consists of 40 H100 nodes, with each compute node equipped with 8 Nvidia H100 GPUs, 2 TB of RAM, and 30 TB of locally attached NVMe storage. 
These nodes are interconnected via a rail-optimised, full fat-tree topology, utilising 400 Gb/s NDR InfiniBand cables.
Additionally, the cluster incorporates a 2.5 PB SSD-based Lustre file system, linked to the H100 nodes through high-speed InfiniBand connections.

With a global batch size of 640, we train the current release of MERaLiON-AudioLLM for around 200k steps, which took 2 days to complete using 128 H100 GPUs.
Prior to this, we had conducted much of our data preprocessing and earlier experiments on other high-performance systems, including NSCC's ASPIRE 2A, Nvidia's Taipei-1 and Europe's LUMI Supercomputer.

\subsection{Tech Stack}
As the field of generative AI continues to evolve rapidly, we often found ourselves grappling with subtle bugs in popular open-source deep learning frameworks such as Huggingface's Transformers\footnote{\url{https://huggingface.co/docs/transformers/en/index}} and PyTorch Lightning.\footnote{\url{https://lightning.ai/docs/pytorch/stable/}}
These challenges are particularly pronounced as we explore the relatively underexplored domain of multimodal architectures.
To avoid spending excessive time addressing bugs and to prioritise our efforts on experimentation, we decided to develop our own training codebase to support the development of MERaLiON-AudioLLM.

Our trainer is implemented primarily in \emph{PyTorch} \citep{paszke2017automatic} and utilises Fully Sharded Data Parallelism (FSDP) to achieve near-linear scaling across all 320 H100 nodes on the ASPIRE 2A+ cluster.
We use \emph{Transformers} to simplify the loading of pre-trained models and \emph{Mosaic StreamingDataset} \citep{mosaicml2022streaming} for efficient large-scale data streaming.
We store our datasets as compressed zstd shards of around 100 MB on the distributed filesystem.
During training, we employ the locally attached NVMe storage as a buffer to prefetch future shards, minimising latency and optimising throughput.
\section{Evaluations}
\label{sec: evaluation}


\subsection{Setup}

We benchmark our model with a series of testsets from AudioBench benchmark~\citep{wang2024audiobench} against three well-known AudioLLMs: Qwen2-Audio 7B \citep{Qwen2-Audio}, WavLLM \citep{hu2024wavllm}, and SALMONN \citep{tang2024salmonn}. 
Qwen2-Audio 7B is a state-of-the-art AudioLLM that combines a Whisper encoder with the Qwen text decoder and outperforms commercial models in audio-centric instruction-following capabilities. WavLLM uses a dual-encoder setup that integrates Whisper with WavLM \citep{chen2022wavlm} and employs a two-stage curriculum learning approach introducing a novel prompt-aware LoRA weight adapter. SALMONN combines a Whisper encoder with BEATS \citep{chen2023beats} and utilises a three-stage curriculum learning strategy for training.
We also compared with an cascaded model, which feeds the transcriptions recognized by Whisper-large-v2 along with the instruction prompts to a Gemma2 9B CPT SEA-LIONv3 Instruct model to get the responses. 
We tuned its hyperparameters and prompt template to optimise performance across a range of speech-to-text tasks.

We evaluate MERaLiON-AudioLLM on 6 tasks, namely automatic speech recognition (ASR), speech translation (ST), spoken question answering (SQA), spoken dialogue summarization (SDS), speech instruction (SI), and paralinguistics (PARA). Evaluation data is derived from the test splits of the following datasets:
\begin{itemize}
    \item ASR: LibriSpeech \citep{librispeech}, Common-Voice-15 \citep{ardila2020common}, Earnings21 \citep{Rio2021Earnings21AP}, Earnings22 \citep{rio2022earnings22}, NSC \citep{koh19_interspeech,wang2025advancing}
    \item ST: CoVoST 2 \citep{wang2021covost}
    \item SQA: SLUE \citep{shon-etal-2023-slue}, Spoken-SQuAD \citep{lee2018spokensquad}, CN-College-Listen-Test \citep{hu2024wavllm}, Singapore-Public-Speech-SQA \citep{wang2024audiobench}, NSC \citep{koh19_interspeech,wang2025advancing}
    \item SDS: NSC \citep{koh19_interspeech,wang2025advancing}
    \item SI: OpenHermes \citep{OpenHermes2.5}, Alpaca \citep{alpaca}
    \item PARA: VoxCeleb \citep{Nagrani_2017}, MELD \citep{poria2019meld}
\end{itemize}
We assess automatic speech recognition (ASR) and speech translation (ST) using Word Error Rate (WER) and BLEU scores \citep{papineni2001bleu}, respectively. For other tasks, we employ the LLM-as-a-Judge framework, which uses a pre-trained large language model to evaluate task performance by generating and scoring responses based on criteria such as relevance, coherence, and accuracy. Data preparation and other implementation details are available in \citet{wang2024audiobench}.

\subsection{Results}

\begin{table*}[ht!]
\scriptsize
\centering
\begin{tabular}{ccccc | c}
\toprule
\textbf{Dataset} & \textbf{MERaLiON} & \textbf{Qwen2-Audio 7B} & \textbf{WavLLM} & \textbf{SALMONN-7B} & \textbf{Cascaded Model}\\ 
\midrule
\emph{Automatic Speech Recognition ($\downarrow$)} &  &  \\   
\cmidrule[0.1pt]{1-1}
LibriSpeech-Test-Clean & 0.03 & 0.03 & \textbf{\underline{0.02}} & 0.10 & 0.03 \\
LibriSpeech-Test-Other & \textbf{\underline{0.05}} & 0.06 & \textbf{\underline{0.05}} & 0.10 & \underline{0.05} \\
Common-Voice-15-En-Test & \textbf{\underline{0.10}} & 0.11 & 0.15 & 0.31 & 0.11 \\
Earnings21-Test & \textbf{0.17} & 0.19 & 0.65 & 0.26 & \underline{0.11} \\
Earnings22-Test & \textbf{0.20} & 0.24 & 0.67 & 0.36 & \underline{0.14} \\
MNSC-ASR-Part 1 & \underline{\textbf{0.05}} & 0.07 & 0.10 & 0.09 & 0.07 \\
MNSC-ASR-Part 2 & \underline{\textbf{0.05}} & 0.19 & 0.45 & 0.42 & 0.33 \\
MNSC-ASR-Part 3 & \underline{\textbf{0.28}} & 0.35 & 0.75 & 0.66 & 0.30 \\
MNSC-ASR-Part 4 & \underline{\textbf{0.40}} & 0.56 & 1.14 & 0.76 & 0.48 \\
MNSC-ASR-Part 5 & \underline{\textbf{0.21}} & 0.28 & 0.40 & 0.35 & 0.23 \\
MNSC-ASR-Part 6 & \underline{\textbf{0.15}} & 0.22 & 0.43 & 0.25 & 0.18 \\

\midrule
\emph{Speech Translation ($\uparrow$)} &  &   \\   
\cmidrule[0.1pt]{1-1}
CoVoST 2 En $\rightarrow$ Id & \textbf{\underline{32.6}} & 16.3 & 13.8 & 14.1 & 27.6 \\
CoVoST 2 En $\rightarrow$ Zh & \textbf{\underline{38.00}} & 25.8 & 32.0 & 33.9 & 35.3 \\
CoVoST 2 En $\rightarrow$ Ta & \textbf{\underline{8.5}} & 0.0 & 0.0 & 0.0 & 8.5 \\
CoVoST 2 Id $\rightarrow$ En & \textbf{37.1} & 6.3 & 5.9 & 26.9 & \underline{46.8} \\
CoVoST 2 Zh $\rightarrow$ En & 15.0 & \textbf{\underline{16.5}} & 2.4 & 5.3 & 15.2 \\
CoVoST 2 Ta $\rightarrow$ En & \textbf{\underline{4.0}} & 0.0 & 0.2 & 0.4 & 2.8 \\

\midrule
\emph{Spoken Question Answering ($\uparrow$)} &  & &  \\   
\cmidrule[0.1pt]{1-1}
SLUE-SQA-5 & 82.9 & 80.1 & \textbf{83.9} & 83.5 & \underline{88.6} \\
Spoken-SQuAD & 70.3 & 64.9 & \textbf{77.7} & 66.4 & \underline{88.6} \\
CN-College-Listen-Test & \textbf{85.0} & 74.5 & 65.4 & 50.9 & \underline{91.9} \\
Singapore-Public-Speech-SQA & \textbf{60.3} & 58.3 & 58.6 & 59.2 & \underline{73.1} \\
MNSC-SQA-Part 3 & \textbf{51.4} & 42.0 & 45.2 & 40.6 & \underline{53.2} \\
MNSC-SQA-Part 4 & \textbf{49.0} & 39.6 & 46.6 & 36.6 & \underline{60.2} \\
MNSC-SQA-Part 5 & \textbf{58.2} & 51.6 & 50.8 & 44.6 & \underline{67.2} \\
MNSC-SQA-Part 6 & \textbf{65.2} & 53.6 & 62.2 & 46.8 & \underline{71.6} \\

\midrule
\emph{Spoken Dialogue Summarization ($\uparrow$)} &  & &  \\   
\cmidrule[0.1pt]{1-1}
MNSC-SDS-Part 3 & \underline{\textbf{46.8}} & 33.8 & 31.6 & 9.0 & 45.4 \\
MNSC-SDS-Part 4 & \underline{\textbf{45.8}} & 24.8 & 31.6 & 7.0 & 44.0 \\
MNSC-SDS-Part 5 & \textbf{55.2} & 40.4 & 45.2 & 17.2 & \underline{58.0} \\
MNSC-SDS-Part 6 & \textbf{61.8} & 46.2 & 49.4 & 24.2 & \underline{65.4} \\

\midrule
\emph{Speech Instruction ($\uparrow$)} & & & \\   
\cmidrule[0.1pt]{1-1}
OpenHermes-Audio & \textbf{71.4} & 44.8 & 22.4 & 15.8 & \underline{72.2} \\
Alpaca-GPT4-Audio & \textbf{73.4} & 52.6 & 21.6 & 17.2 & \underline{73.8} \\

\midrule
\emph{Paralinguistics ($\uparrow$)} &  & &  \\   
\cmidrule[0.1pt]{1-1}
VoxCeleb-Gender-Test & \textbf{\underline{99.5}} & 99.1 & 69.7 & 88.8 & 35.3 \\
VoxCeleb-Accent-Test & \textbf{\underline{46.4}} & 29.2 & 35.8 & 34.2 & 24.6 \\
MELD-Sentiment-Test & 42.3 & \textbf{53.5} & 50.1 & 42.1 & \underline{56.7} \\
MELD-Emotion-Test & 30.2 & 40.5 & \textbf{41.1} & 30.7 & \underline{47.4} \\



\bottomrule
\end{tabular}
\caption{Evaluation Results on AudioBench Datasets: The best result for each dataset is underlined, while the top-performing AudioLLM result is highlighted in bold.}

\label{tab:eval_results}
    
\end{table*}

As shown in Table \ref{tab:eval_results}, MERaLiON-AudioLLM is competitive against other AudioLLMs on many datasets.
For instance, our model has the best performance on the unseen Earnings21-Test and Earning22-Test datasets, beating the second-best AudioLLM by significant margin.
As expected, AudioLLM performs better on the NSC datasets, given its training on in-domain data. This is especially evident in the MNSC ASR Part 2 dataset, which contains prompted readings with references to people, food, locations, and brands associated with Singapore. In this context, MERaLiON-AudioLLM significantly outperforms both Qwen2-Audio (19\% word error rate) and Whisper-large-v2 (33\% word error rate), achieving an impressive word error rate of just 5\%.
These results demonstrate that while strong open-sourced models like Whisper are effective, they still require localisation to optimise performance in specific, localised contexts.
On the other hand, we observe that AudioLLMs are prone to the ``diverse prompts'' issue \citep{wang2024audiobench}, where performance on standard benchmarks like LibriSpeech slightly declines when exposed to a set of unseen text prompts. This leads to higher word error rates compared to traditional ASR models, such as Whisper.

The only tasks that our model clearly underperforms other models are MELD-Sentiment-Test and MELD-Emotion-Test, where the goal is to identity the sentiment or the emotion of the speaker based on the speech.
To address this, we are actively curating additional paralinguistic datasets that focus on local contexts and accents. Additionally, we are exploring architectural improvements to better integrate paralinguistic information into our model. 
\section{Related Work}

\subsection{AudioLLMs}

Recent advancements in large language models (LLMs) have paved the way for their integration with speech and audio processing capabilities, enabling the development of end-to-end systems that efficiently streamline the workflow from audio input to meaningful output. Some of these integrations focus on specific tasks, such as automatic speech recognition and speech translation \citep{Yu2023ConnectingSE, ma2024embarrasingly, xu2024compare, chen2024llast, wu2023speechllama}. These specialized models leverage features extracted by speech encoders, such as Whisper \citep{radford2023whisper} and Hubert \citep{hsu2021hubert}, to enhance their understanding and processing of spoken language. Beyond these specialized tasks, there is a growing trend towards multitask models capable of performing multiple audio-related tasks within a unified framework.
Audio Flamingo \citep{Kong2024AudioFlamingo}, Pengi \citep{deshmukh2023pengi}, and GAMA \citep{ghosh2024GAMA} focus on non-speech audio tasks such as audio captioning, audio question answering, and understanding non-verbal speech cues. More generally, AudioLLMs handle a diverse range of speech and non-speech audio tasks. For instance, SALMONN \citep{tang2024salmonn} is evaluated on 15 tasks in 3 levels of difficulty from in-distribution tasks seen at training to new out-of-distribution tasks. A common approach is to connect one or more audio encoders to language models to facilitate the processing of audio inputs and the generation of text-based outputs \citep{tang2024salmonn, hu2024wavllm, gong2024ltu, gong2023ltuas, fathullah2024AudioChatLlama, chu2023qwenaudio, Qwen2-Audio, xie2024MiniOmni, chen2024bestow, held2024diva, nguyen2024spiritlm}. The encoder component is typically initialized with off-the-shelf audio models trained either in supervised or self-supervised learning paradigms \citep{radford2023whisper, hsu2021hubert, chen2023beats} to extract semantic and acoustic features. The language model component is typically initialized with off-the-shelf decoder-only LLMs \citep{llama3modelcard, Touvron2023Llama2, chen2022wavlm, bai2023qwen}. These pre-trained models provide a robust foundation for feature extraction and modeling to enhance the AudioLLM’s performance on downstream tasks. A large number of existing works use continuous audio embeddings, and others use discrete audio embeddings by quantizing the continuous embeddings or through neural audio codecs \citep{rubenstein2023audiopalm, wang2023viola, an2024funaudiollm, defossez2024moshi, fang2024LlamaOmni}. In the latter, the vocabulary size is expanded with the audio tokens and the AudioLLM can be trained to directly consume and output text and audio tokens.

\subsection{Multimodal Fusion}

Traditional cascaded systems, which combine a speech recognizer with a large language model, are commonly employed in multimodal applications due to their simplicity and modularity. However, they may face limitations in responsiveness and capturing complex contextual information. End-to-end solutions require the model to learn the relationships between audio and text in a more integrated manner. 

Some works have explored architectural and training strategies. Audio Flamingo \citep{Kong2024AudioFlamingo} employs audio representation transformation layers to condition on audio inputs. 
\citet{Yu2023ConnectingSE} studies the effectiveness of different modality adapter modules.
\citet{ma2024embarrasingly} trains only the randomly initialized adapter for ASR tasks, and other works such as LLaMA-Omni \citep{fang2024LlamaOmni} and SpeechGPT \citep{zhang2023speechgpt} instruction finetune additional model components with curated datasets. WavLLM \citep{hu2024wavllm} uses a curriculum learning framework to first build foundational capabilities through learning elementary tasks, then build more sophisticated capabilities by learning more complex tasks that combine elementary tasks. SALMONN \citep{tang2024salmonn} employs activation tuning at the end of a multi-stage training scheme to help regain the emergent abilities of LLMs and mitigate catastrophic forgetting.

A key goal in multimodal fusion is to align both audio and text into a common representational space. SpeechGPT \citep{zhang2023speechgpt} uses cross-modal instruction finetuning to align the inputs and outputs of the two different modalities, and proposes chain-of-modality instruction finetuning to map speech input into text representation space, think about the answering process in text, and then output speech or text response. AudioChatLlama \citep{fathullah2024AudioChatLlama} and DiVA \citep{held2024diva} distill the responses of text-only LLMs to speech transcripts. Spirit LM \citep{nguyen2024spiritlm} seeks word-level alignment by interleaving speech and text in a single sequence.

Besides audio-text alignment, acoustic features capturing paralinguistic information such as tone, pitch and emotion can be incorporated during fusion to provide context beyond the words themselves. DeSTA \citep{lu2024desta} proposes a descriptive speech-text alignment approach by leveraging paralinguistic metadata to generate captions with natural language descriptions of the acoustic quality of speech inputs, so as to capture both the spoken words and speaking styles. SpeechEmotionLlama \citep{kang2024speechemotionllama} distills the responses of LLMs to speech transcripts where the speaker’s emotion is specified. Some works use additional encoders to extract acoustic features \citep{tang2024salmonn, hu2024wavllm, zhang2024mowe}. In Moshi \citep{defossez2024moshi}, the neural audio codec encodes audio into semantic tokens that capture linguistic content, and acoustic tokens that are optimized for high-quality audio reconstruction to retain fine audio details. EMOVA \citep{chen2024emova} also disentangles the semantic and acoustic features of speech inputs, which subsequently allows changing the style of speech while preserving its semantic content.

\section{Limitations and Future Work}


Although MERaLiON-AudioLLM demonstrates competitive performances on standard benchmark evaluations, there are several limitations that we are aware of and would address in future releases:

\paragraph{Safety}
As MERaLiON-AudioLLM have not been specifically aligned for safety and may generate content that is inappropriate, offensive, or harmful. Developers and users are responsible for performing their own safety fine-tuning and implementing necessary security measures. The authors shall not be held liable for any claims, damages, or other liabilities arising from the use of the released models, weights, or code.

\paragraph{Context Length}
Currently, MERaLiON-AudioLLM supports only up to 30 seconds of audio context. We are actively working on improving its ability to process and integrate long-range dependencies in conversational speech and complex narratives. Additionally, we are enhancing the model's capacity to handle multi-turn interactions and interleaved text and audio inputs.

\paragraph{Loss of instruction following capability}
Achieving strong performance in tasks like speech recognition and speech translation requires fine-tuning the entire AudioLLM in an end-to-end manner, but this has resulted in some degree of catastrophic forgetting, causing the model to lose the ability to follow certain text instructions. 
To address this, we have planned several key explorations before the next release. 
First, we plan to create a Speech Instruction Following dataset to measure the extent of this capability loss. Additionally, we will experiment with training the AudioLLM on more diverse multimodal datasets while maintaining a replay collection of text-only instruction pairs to mitigate forgetting. Finally, we are exploring methods to directly align audio encoders with text decoders, bypassing the need to fine-tune the text decoders. Early successes in this area show promise, and we intent to explore this line of work further.

\paragraph{Multilingualism and Empathetic Reasoning} While MERaLiON demonstrates the ability to handle non-English speech-text tasks, as well as non-speech tasks such as Emotion and Gender Recognition, we believe its performance on these tasks can be further enhanced with additional data. To that end, we are actively exploring strategies to scale up our data collection efforts in an efficient manner.




\section{Conclusion}
This technical report presents MERaLiON-AudioLLM, a multimodal model designed to bridge the gap between speech and text. 
Through careful data curation and optimisation, MERaLiON-AudioLLM demonstrates significant advancements in both speech-text understanding and generation, especially in local contexts.
The development of MERaLiON-AudioLLM highlights the potential of combining large-scale multimodal datasets with advanced model architectures to address real-world challenges. 
However, there remains areas for further exploration, particularly in refining instruction-following behaviour and enhancing robustness across low-resource datasets.
We look forward to continuing this line of research, contributing to the advancement of multimodal AI systems, and making our work available to the broader research community for further validation and development.

\section*{Acknowledgement}

    We extend our sincere gratitude to Jeremy H. M. Wong, Kye Min Tan, Geyu Lin, Ziyi Xu, Hardik B. Sailor, Qiongqiong Wang, Tianchi Liu, Muhammad Huzaifah, Xin Huang, Tarun K. Vangani, Nattadaporn Lertcheva, Xi Wang, Kui Wu, Siti Maryam Binte Ahmad Subaidi, Madalene Hee for their invaluable contributions to data management, human annotations, logistics, insightful discussions, and future work explorations. 
    
    This research is supported by the National Research Foundation, Singapore and Infocomm Media Development Authority, Singapore under its National Large Language Models
    Funding Initiative. Any opinions, findings and conclusions or
    recommendations expressed in this material are those of the
    author(s) and do not reflect the views of National Research
    Foundation, Singapore and Infocomm Media Development
    Authority, Singapore.

\bibliographystyle{abbrvnat}
\bibliography{anthology,custom}

\end{document}